\documentclass{article}

\usepackage{arxiv}

\usepackage[utf8]{inputenc} 
\usepackage[T1]{fontenc}    
\usepackage{hyperref}       
\usepackage{url}            
\usepackage{booktabs}       
\usepackage{amsfonts}       
\usepackage{amsmath}
\usepackage{nicefrac}       
\usepackage{microtype}      
\usepackage{cleveref}       
\usepackage{graphicx}
\usepackage{xcolor}
\usepackage{multirow}
\usepackage{subcaption}
\usepackage{natbib}
\usepackage{doi}

\usepackage{amsmath,amsfonts,bm}

\def\eqref#1{equation~\ref{#1}}

\def\1{\bm{1}}

\DeclareMathAlphabet{\mathsfit}{\encodingdefault}{\sfdefault}{m}{sl}
\SetMathAlphabet{\mathsfit}{bold}{\encodingdefault}{\sfdefault}{bx}{n}

\title{EvoAD: Self-Evolving Autonomous Driving via Latent World-Model Guided Post-hoc Reflection}
\date{}

\author{
  {\bfseries Junjie Yang\textsuperscript{1} \quad Qiming Zhang\textsuperscript{2} \quad Yuyan Li\textsuperscript{2} \quad Shuqi Shen\textsuperscript{3} \quad Yusong Huang\textsuperscript{2}} \\[0.5em]
  {\bfseries Yanggang Sheng\textsuperscript{2} \quad Wenpeng Xu\textsuperscript{2} \quad Ziyi Shi\textsuperscript{1} \quad Tianshun Li\textsuperscript{2} \quad Yijie Chen\textsuperscript{2}} \\[0.5em]
  {\bfseries Yujia Wang\textsuperscript{4} \quad Jun Ma\textsuperscript{2} \quad Xinhu Zheng\textsuperscript{2} \quad Hongliang Lu\textsuperscript{5} \quad Hai Yang\textsuperscript{1}}
}

\renewcommand{\shorttitle}{EvoAD: Self-Evolving Autonomous Driving}
\renewcommand{\headeright}{}
\renewcommand{\undertitle}{}
\hypersetup{
pdftitle={EvoAD: Self-Evolving Autonomous Driving via Latent World-Model Guided Post-hoc Reflection},
pdfsubject={cs.RO, cs.AI},
pdfauthor={Junjie Yang, Qiming Zhang, Yuyan Li, Shuqi Shen, Yusong Huang, Yanggang Sheng, Wenpeng Xu, Ziyi Shi, Tianshun Li, Yijie Chen, Yujia Wang, Jun Ma, Xinhu Zheng, Hongliang Lu, Hai Yang},
pdfkeywords={Autonomous Driving, World Model, Self-Evolving, Post-hoc Reflection},
}

\begin{document}
\maketitle

\begingroup
\let\thefootnote\relax
\footnotetext{\textsuperscript{1}\,The Hong Kong University of Science and Technology}
\footnotetext{\textsuperscript{2}\,The Hong Kong University of Science and Technology (Guangzhou)}
\footnotetext{\textsuperscript{3}\,The Chinese University of Hong Kong, Shenzhen}
\footnotetext{\textsuperscript{4}\,Victoria University of Wellington}
\footnotetext{\textsuperscript{5}\,Southern University of Science and Technology}
\endgroup

\begin{abstract}
  
  End-to-end autonomous driving systems have achieved rapid progress by learning unified policies that map sensor observations directly to driving trajectories. However, once deployed, these systems typically remain frozen: they may repeat similar decisions in similar scenes, and failures can only be addressed through costly data collection and retraining. This sharply contrasts with human drivers, who improve lifelong through experience, memory, and reflection. In this work, we ask whether autonomous driving systems can achieve similar post-deployment self-evolution to human drivers. We propose EvoAD, a self-evolving autonomous driving framework via latent world-model guided post-hoc reflection. EvoAD stores driving experiences in a bird's-eye-view feature memory pool, retrieves scene-relevant precedents according to geometric and semantic similarity, and uses a latent world model to evaluate the future consequences of candidate trajectories in a compact planning-oriented space. Instead of treating memory as external textual contexts or using world models only for pre-action simulation, EvoAD transforms executed driving outcomes into reusable improvement signals for future decisions. This enables the system to diagnose how prior trajectories could have been improved and to refine behavior in similar future scenarios without retraining the base driving policy. Experiments on NAVSIM v1 and v2 show consistent improvements across reflection rounds and state-of-the-art performance, demonstrating the potential of world models as reflective mechanisms for post-deployment self-evolution in autonomous driving.
  
\end{abstract}

\section{Introduction}


End-to-end (E2E) autonomous driving (AD) has advanced rapidly as a unified data-driven paradigm, where a single neural policy is trained on large-scale driving data to map raw sensor observations directly to driving trajectories~\citep{hu2023planning,jiang2023vad}.
Yet after training, these systems are typically deployed as frozen models: given the same scene, they are likely to repeat the same decision, and any failure observed after deployment can only be corrected through another costly cycle of data collection, labeling, and retraining~\citep{chen2024endtoend, caesar2021nuplan}. In stark contrast, human drivers exhibit a fundamentally different form of competence. Even after obtaining a license, they continue to improve lifelong: they remember previously encountered situations, mentally revisit decisions that felt unsafe or suboptimal, and adjust their future behavior when similar scenarios arise~\citep{groeger2000driving, fuller2005driverbehaviour}. This contrast exposes a central limitation of current E2E AD systems: driving competence is never achieved once and for all; it continually evolves through experience, memory, and reflection.

\begin{figure}[t]
    \centering
    \includegraphics[
        width=0.80\linewidth
    ]{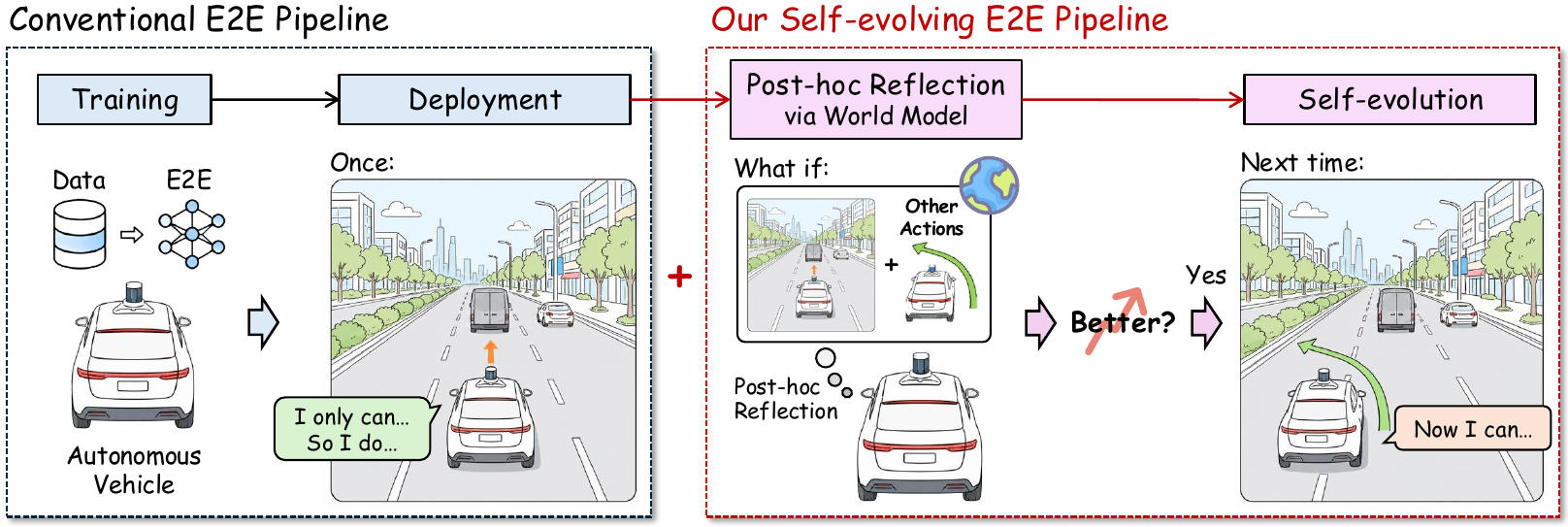}
    \caption{\textbf{From Conventional to Self-Evolving E2E AD}. Conventional E2E pipelines typically execute fixed policies after deployment and thus tend to repeat similar decisions in similar scenarios. We introduce a world-model guided post-hoc reflection mechanism that enables the system to imagine alternative actions, identify potentially better outcomes, and update future behavior.}
    \label{fig:teaser}
    \vspace{-1.5em}
\end{figure}


A growing line of work has sought to recover this missing adaptability by moving self-improvement into the test-time loop, including online or test-time fine-tuning~\citep{wang2021tent,hanselmann2022king}, knowledge- and retrieval-augmented driving~\citep{yuan2024ragdriver, wen2023dilu}, and score-based candidate re-ranking~\citep{li2024hydramdp,gtrs2025}. However, retrieval and memory in these methods are typically used as external context, such as textual knowledge, high-level experience, or demonstration priors. They rarely provide feedback grounded in the actual consequences of the trajectory just produced by the planner. More recently, world models promise to bridge this gap by creating an imaginative space for action evaluation, where each candidate trajectory can be rolled forward into its possible visual or latent consequences~\citep{wang2024drivedreamer,gao2024magicdrive, lu2024wovogen}. Consequently, many AD approaches have incorporated world models as forward simulators, using them to imagine future outcomes before selecting an action, such as World4Drive~\citep{zheng2025world4drive}, and Drive-WM~\citep{wang2024drivewm}. In these systems, world models behave like a forward-looking advisor: they imagine possible futures to guide action selection before the vehicle acts. This dominant use of world models, however, obscures a broader question about their role in AD: can their imaginative capacity support human-like self-evolution through post-hoc reflection?

In this study, we propose \textbf{EvoAD}, a self-evolving AD framework based on latent world-model guided post-hoc reflection. As shown in Fig.~\ref{fig:teaser}, rather than treating a trained E2E model as a fixed policy, EvoAD converts past driving experiences into reusable improvement signals for future decision-making. 
First, we introduce bird's-eye-view (BEV) feature-based memory to store reflected driving experiences, including scene representations, trajectories, future latents, and evaluation feedback. Compared with text-only memory or high-level action descriptions, BEV-level memory better preserves the geometric layout and spatial relations of the driving scene, making retrieval more aligned with the structure of AD decisions. 
Second, we use a latent world model to imagine the future consequences of candidate trajectories in a compact planning-oriented latent space, which enables safer or more efficient trajectory evaluation without expensive pixel-level future reconstruction. 
Third, EvoAD performs outcome-aware post-hoc reflection based on trajectory evaluation feedback, so the reflection is not merely past-action evidence, but a targeted improvement based on the previous weakness. In doing so, EvoAD enables a trained base model to accumulate experience, retrieve relevant prior reflections, and refine future behavior without retraining.

In summary, our contributions are threefold: 1) We propose a self-evolving AD framework based on post-hoc reflection, enabling AD systems to improve after deployment without retraining the base model; 2) We redefine the role of the world model as a post-hoc reflective evaluator, which explores better alternatives in a latent imaginative space; and 3) We introduce BEV-feature-based memory and retrieval, which stores scenario-level driving experiences and retrieves relevant ones to support future decision refinement. Experiments on NAVSIM v1 and v2 show that EvoAD improves WoTE from 88.3 to 93.4 PDMS and from 84.2 to 91.5 EPDMS, respectively, achieving state-of-the-art performance on both benchmarks.

\section{Related Work}

\subsection{Post-hoc Reflection and Retrieval Memory for Autonomous Driving}

Recent vision-language and multimodal driving models increasingly use reasoning, explanation, retrieval, and memory to support driving decisions. Existing studies have explored scene-graph-based explanations for auditable decision making~\citep{belmecheri2024trustworthy}, retrieval-augmented expert demonstrations for control and explanation~\citep{yuan2024ragdriver}, reward-aligned reasoning for trajectory or meta-action prediction~\citep{li2025driver1}, and diagnostic analyses of the reasoning-planning disconnect in driving vision-language models. In parallel, retrieval and memory have been introduced to incorporate traffic regulations, expert trajectories, commonsense knowledge, and past experience into driving systems~\citep{cai2024drivereg,fu2025orion,ye2025safedriverag}. Yet prior methods mainly use post-hoc reasoning for explanation or retrieval as auxiliary knowledge, external demonstrations, or planner priors, lacking a closed-loop mechanism that reflects on the model’s own driving outcomes and turns them into reusable improvement signals. EvoAD fills this gap by using latent world-model feedback to evaluate its own trajectory consequences, store reflected experiences in BEV-feature memory, and retrieve relevant experiences to refine future decisions.

\subsection{Latent World Modeling in Autonomous Driving}

World models provide a natural substrate for deliberative AD by simulating possible future scene evolutions before action execution. Early driving world models such as Drive-WM couple multiview future visual forecasting with planning by rolling out candidate futures and selecting actions according to predicted outcomes~\citep{wang2024drivewm}. Recent methods increasingly move from pixel-level generation to planning-oriented latent representations: LAW predicts future scene features as self-supervision for E2E driving~\citep{li2025law}; World4Drive learns intention-aware physical latent dynamics for multimodal trajectory selection and closed-loop performance~\citep{zheng2025world4drive}; WorldRFT fine-tunes latent world models with reinforcement signals to better align them with planning objectives~\citep{wang2025worldrft}; and DriveWorld-VLA unifies vision-language-action planning and latent-space world modeling within a shared latent state~\citep{jia2026driveworldvla}. These studies establish latent world models as planning-aware future simulators. EvoAD builds on this trend but uses the latent world model differently: instead of merely predicting future states or selecting actions, it uses latent future feedback for post-hoc reflection and iterative self-evolution of the driving policy.

\section{Methodology}
\begin{figure}[t]
    \centering
    \includegraphics[
        width=0.98\linewidth,
    ]{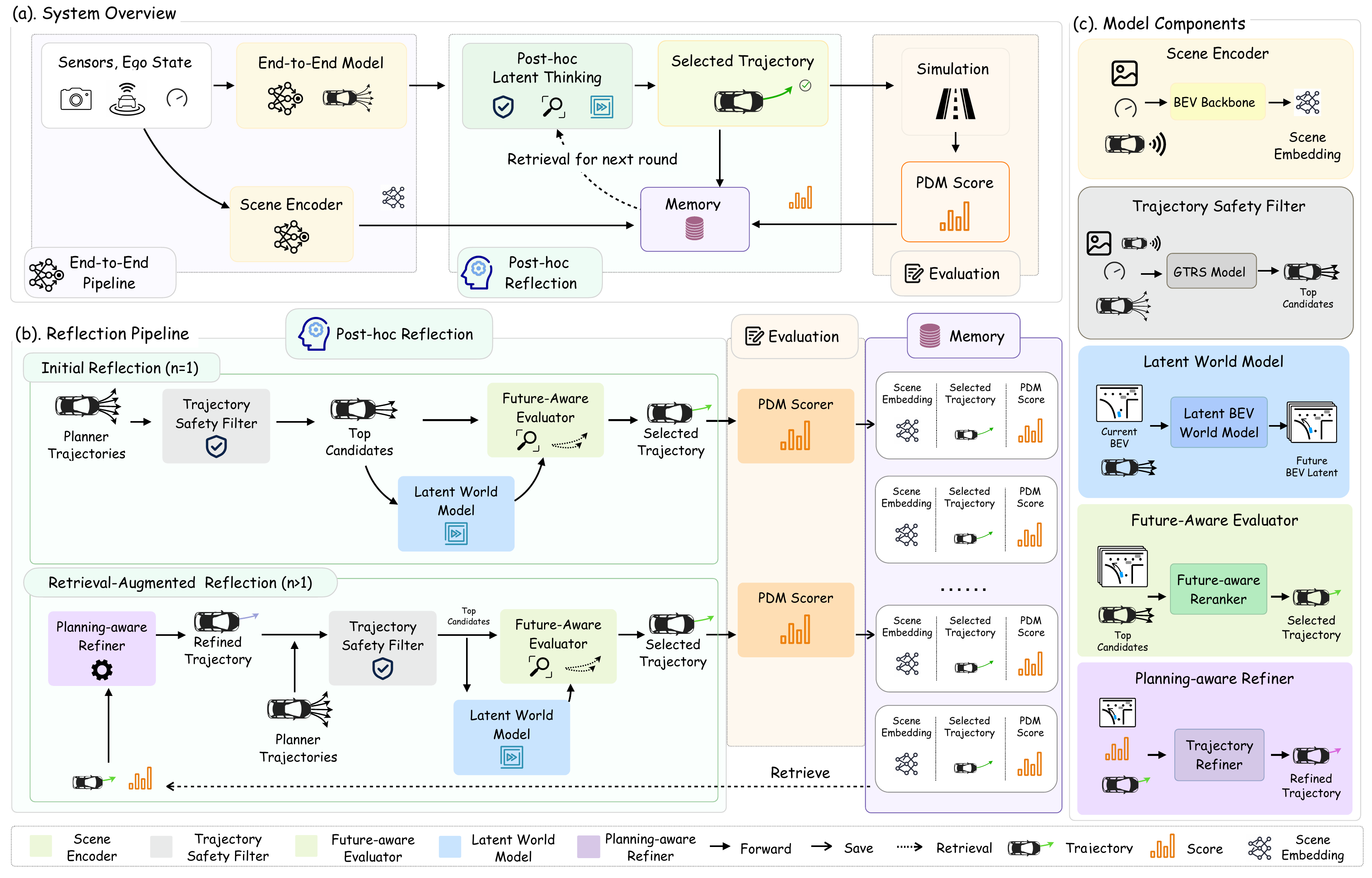}
    \caption{
        \textbf{Overview of EvoAD.}
(a) \textbf{System Overview}: an E2E planner augmented with post-hoc latent reflection, evaluation, and memory. (b) \textbf{Reflection Pipeline}: safe, future-aware trajectories are selected and then refined with retrieved experiences. (c) \textbf{Model Components}: scene encoder, safety filter (SF), latent world model with future-aware evaluator (FAE), and planning-aware refiner (PAR).
    }
    \label{fig:evo-f1}
    \vspace{-1em}
\end{figure}

\subsection{Problem Formulation}\label{sec:problem_formulation}

We formulate post-hoc trajectory improvement for E2E AD as an iterative process followed by E2E model inference and evaluation results.
At each timestamp $t$, the system receives a driving observation
$\mathcal{O}_t=\{\mathcal{I}_t,\mathbf{s}_t^{\mathrm{raw}},\mathcal{G}_t,c_t\}$,
where $\mathcal{I}_t$ denotes sensor observations, $\mathbf{s}_t^{\mathrm{raw}}$ is the
raw ego state, $\mathcal{G}_t$ represents map context, and $c_t$ is the navigation
command. Given $\mathcal{O}_t$, a base planner $\pi$ generates a set of candidate
future trajectories
\begin{equation}
    \mathcal{Y}_t^{(1)}=\pi(\mathcal{O}_t)
    =\{\mathbf{y}_{t,k}^{(1)}\}_{k=1}^{K}.
\end{equation}
Instead of replacing the base planner, EvoAD aims to improve its output through
a multi-round post-hoc reflection process,
\begin{equation}
    \mathbf{y}_t^\star
    =
    \mathcal{F}(\mathcal{O}_t,\mathcal{Y}_t^{(1)},\mathcal{M}),
\end{equation}
where $\mathcal{F}$ denotes the proposed reflection procedure and $\mathcal{M}$ is
an episodic memory that stores previous scene-level planning experiences.

As shown in Fig.~\ref{fig:evo-f1}, each reflection round chains a reusable SF, an FAE backed by a latent world model, and a post-hoc feedback signal on the executed trajectory; from the second round onward, the PAR additionally consumes memory-retrieved
experiences to correct the previously selected trajectory before re-evaluation.
The whole loop can be abstracted as
\begin{equation}
    \mathbf{y}_t^{(n)}, r_t^{(n)}, \mathcal{M}
    = \mathcal{R}\!\left(
        \mathcal{O}_t,\,\mathcal{Y}_t^{(n)},\,\mathcal{M}
    \right),
    \qquad n=1,\dots,N,
\end{equation}
where $\mathcal{R}$ denotes one reflection step, $\mathbf{y}_t^{(n)}$ the selected
trajectory at round $n$, and $r_t^{(n)}$ its post-hoc feedback;
$\mathbf{y}_t^\star = \mathbf{y}_t^{(N)}$ is returned after $N$ rounds. Here, $r_t^{(n)}$ is observed only after execution and is never used for selection. It can be derived from real-world feedback signals, such as takeovers, jerk, or lane deviation; in our experiments, we use NAVSIM's PDMS, a composite metric capturing safety, progress, and comfort.

\subsection{Future-Latent Guided Post-hoc Reflection}
\label{sec:future_latent_reflection}
A trajectory that appears feasible under the current observation may still lead to unsafe
or inefficient future interactions. EvoAD therefore augments post-hoc trajectory
evaluation with candidate-conditioned future latents from a latent world model. At
reflection round $n$, the score-based SF first prunes evidently
infeasible candidates from $\mathcal{Y}_t^{(n)}$, resulting in
$\tilde{\mathcal{Y}}_t^{(n)}$. For each remaining candidate
$\mathbf{y}_{t,k}^{(n)}$, we obtain a future latent representation
\begin{equation}
    \hat{\mathbf{z}}_{t,k}^{(n)}
    =
    \textsc{LatentWM}
    \left(
    \mathcal{O}_t,
    \mathbf{y}_{t,k}^{(n)}
    \right),
\end{equation}
which summarizes the predicted future scene evolution conditioned on executing that
candidate trajectory. EvoAD does not directly decode this latent representation or use the world model as a planner. Instead, the future latent is used as a compact trajectory-conditioned prior to provide the future trend for downstream evaluation. We form a future-latent augmented candidate representation:
\begin{equation}
    \mathbf{h}_{t,k}^{(n)}
    =
    \operatorname{Concat} 
    \left(
    \mathbf{s}_t,
    \mathbf{y}_{t,k}^{(n)},
    \hat{\mathbf{z}}_{t,k}^{(n)}
    \right),
\end{equation}
where $\mathbf{h}_{t,k}^{(n)}$ is the concatenated feature of the ego status, the candidate trajectory, and its predicted future latent. This representation allows the evaluator to compare candidates not only by their geometric feasibility, but also by their potential future consequences from latent features. The same future-latent interface is used for both initial planner candidates and memory-refined proposals, providing a unified bridge between the latent world model and iterative post-hoc reflection.

\subsection{Future-Aware Trajectory Evaluator}
\label{sec:evaluator}

Since score-based methods like GTRS \citep{gtrs2025} only evaluate the performance of trajectories under current scene feature without considering the potential future latent signal, we introduce the FAE to rerank the candidate proposals.
In each reflection round, the FAE scores the safety-filtered
candidates and selects the one EvoAD will execute. Building on
$\mathbf{h}_{t,k}^{(n)}$ from Sec.~\ref{sec:future_latent_reflection}, it (i)
encodes each candidate into a compact feature, (ii) jointly grounds the
compact feature in the current scene and compares it against the other candidates, and
(iii) decodes a ranking signal by dense score supervision of each candidate.

\paragraph{Intra-candidate encoding.}
For each candidate trajectory $\mathbf{y}_{t,k}^{(n)}$, the evaluator receives its
future-latent augmented representation $\mathbf{h}_{t,k}^{(n)}$, then encoded into a compact feature through a self-attention module, which summarizes the future evidence of a single proposal:
\begin{equation}
    \mathbf{c}_{t,k}^{(n)}
    =
    \operatorname{SelfAttn}
    \left(
    \mathbf{h}_{t,k}^{(n)}
    \right).
\end{equation}

\paragraph{Current context fusion.} 
To further relate this future-aware proposal feature $\mathbf{c}_{t,k}^{(n)}$ to the
current driving scene, we reuse the backbone scene feature
$\mathbf{F}_t\in\mathbb{R}^{L\times d}$ as a shared scene context and update each
candidate feature through cross-attention:
\begin{equation}
    \mathbf{c}_{t,k}^{\mathrm{ctx},(n)}
    =
    \operatorname{CrossAttn}
    \left(
    \mathbf{c}_{t,k}^{(n)}, \mathbf{F}_t
    \right)+\alpha \mathbf{c}_{t,k}^{(n)}.
\end{equation}
This operation allows each future-aware proposal feature to query the current scene evidence, so that the model can incorporate predicted future evolution while preserving an explicit understanding of the present traffic context. We apply this query update with a residual connection
and a learnable gate $\alpha$, so that the evaluator can adaptively balance the original
future-aware candidate representation and the current-scene contextual feature. 

\paragraph{Set-wise candidate interaction.} Considering the fact that proposal quality is relative within the same
scene, the evaluator should compare candidates under shared context rather than score
each trajectory independently. After context fusion, all candidates are jointly processed by set-wise self-attention to interact with each other for further ranking:
\begin{equation}
    [\hat{\mathbf{c}}_{t,1}^{(n)},\dots,\hat{\mathbf{c}}_{t,K}^{(n)}]
    =
    \operatorname{SelfAttn}
    \left(
    [\mathbf{c}_{t,1}^{\mathrm{ctx},(n)},\dots,
     \mathbf{c}_{t,K}^{\mathrm{ctx},(n)}]
    \right).
\end{equation}
The set-wise interaction therefore helps capture
redundancy, relative safety, progress trade-offs, and fine-grained differences among
similar proposals.


\paragraph{Training loss.}
Given the candidate feature $\hat{\mathbf{c}}_{t,k}^{(n)}$, the evaluator
uses a ranking head to decode the ranking signal $\hat{s}_{t,k}^{(n)}$ and auxiliary score heads to predict trajectory score and sub-items $\hat{r}_{t,k}^{(n)}$, $\hat{\mathbf{u}}_{t,k}^{(n)}$:
\begin{equation}
    \hat{s}_{t,k}^{(n)} = h^{\mathrm{rank}}(\hat{\mathbf{c}}_{t,k}^{(n)}),
    \qquad
    \mathbf{\hat{u}}_{t,k}^{(n)} = h^{\mathrm{sub}}(\hat{\mathbf{c}}_{t,k}^{(n)}),
    \qquad
    \hat{r}_{t,k}^{(n)} = h^{\mathrm{score}}(\hat{\mathbf{c}}_{t,k}^{(n)}).
\end{equation}
The ranking score $\hat{s}_{t,k}^{(n)}$ is used to select the final proposal. Then, the evaluator is trained jointly with
\begin{equation}
    \mathcal{L}_{\mathrm{eval}}
    = \lambda_{\mathrm{list}}\,\mathcal{L}_{\mathrm{list}}
    + \lambda_{\mathrm{sub}}\,\mathcal{L}_{\mathrm{sub}}
    + \lambda_{\mathrm{score}}\,\mathcal{L}_{\mathrm{score}},
\end{equation}
where $\mathcal{L}_{\mathrm{list}}$ optimizes the within-scene proposal ordering on
$\hat{s}_{t,k}^{(n)}$ by the maximum-likelihood pairwise loss, $\mathcal{L}_{\mathrm{score}}$ and $\mathcal{L}_{\mathrm{sub}}$ denote the loss of predicted scores \citep{niu2025ipad}.

\subsection{Retrieval-Guided Planning-aware Refiner}

\label{sec:refiner}
From Round 2 onward, EvoAD looks up similar past experiences
in an episodic memory and uses them to correct the previously selected trajectory. With the last-round decision and its performance retrieved from the memory, 
the refiner aims to consume the warm-start trajectory together with its real post-hoc feedback
and per-trajectory future latent, and then emit a small, feedback-conditioned
correction rather than re-planning. By targeting the failure mode observed in the
previous round (dangerous interaction, low progress, out-of-drivable-area, etc.), the
refinement is specific for trajectory deficiencies rather than searching for a higher score.

\paragraph{Episodic memory.}
EvoAD maintains an episodic memory $\mathcal{M}$ that accumulates evaluated
planning experiences. Each entry stores the round's selected candidate as
\begin{equation}
    m_{t,k}^{(n)} = \bigl(\mathbf{e}_t,\, \mathbf{y}_{t,k}^{(n)},\, r_{t,k}^{(n)},\,
                          \mathbf{u}_{t,k}^{(n)},\, \hat{\mathbf{z}}_{t,k}^{(n)}\bigr),
\end{equation}
where $\mathbf{e}_t$ is the scene embedding from the scene encoder,
$\mathbf{y}_{t,k}^{(n)}$ the selected trajectory, $(r_{t,k}^{(n)},
\mathbf{u}_{t,k}^{(n)})$ its measured feedback total and sub-scores, and $\hat{\mathbf{z}}_{t,k}^{(n)}$ its per-trajectory future latent.
Each scene keeps at most one record, overwritten only on improvement, so the memory does not grow with the number of rounds. Nor does it need to grow with the number of scenes: once the pool is sufficiently rich it can be frozen, or even built entirely from another domain (Sec.~\ref{sec:cross_domain}). Given $\mathbf{e}_t$, the related scenes are retrieved by cosine similarity,
\begin{equation}
    \mathcal{N}_t^{(n)}
    = \operatorname{TopK}_{m_{t',k'}^{(n')}\in\mathcal{M}}
      \operatorname{sim}\bigl(\mathbf{e}_t,\mathbf{e}_{t'}\bigr),
\end{equation}
yielding scene-conditioned priors over previously observed planning patterns.

\paragraph{Planning-aware refinement model.}

Let $\mathbf{y}_t^{w}$ denote the last-round selected trajectory retrieved from the memory $\mathcal{N}_t^{(n)}$ to be refined, and $k_w$ is the search index. Our planning-aware
refinement model takes $\mathbf{y}_t^{w}$, the per-trajectory future feature
$\hat{\mathbf{z}}_{t,k_w}^{(n)}$, the ego-status feature $\mathbf{s}_t$, and the planning feedback vector
$\mathbf{q}_t=[r_t,\mathbf{u}_t]$, where $r_t$ and $\mathbf{u}_t$ are the measured post-hoc total and sub-score feedback. Rather than generating a new plan
from scratch, the model is designed to perform a conservative, feedback-driven
correction of the selected proposal.

The trajectory, scene feature, ego-status feature, and planning feedback are first
embedded into four tokens and fused by a lightweight self-attention block $H$. Then we inject the planning feedback directly into the trajectory
representation $\mathbf{t}_{H}^{\mathrm{traj}}$ through FiLM~\citep{perez2018film} modulation, so that the refinement is explicitly
conditioned on why the current trajectory is weak:
\begin{equation}
    [\boldsymbol{\gamma},\boldsymbol{\beta}]
    = \psi_{\mathrm{FiLM}}(\mathbf{q}_t),
    \qquad
    \mathbf{t}'_t
    = (1+\boldsymbol{\gamma})\odot \mathbf{t}_{H}^{\mathrm{traj}}
    + \boldsymbol{\beta},
\end{equation}
where $\mathbf{t}_{H}^{\mathrm{traj}}$ denotes the trajectory embedding after attention fusion. The modulated
feature set is then decoded through MLPs to predict both a residual trajectory correction $\Delta\mathbf{y}_t$ and a learnable confidence gate $g_t$, and the refined trajectory is obtained by a gated residual update,
\begin{equation}
    \bar{\mathbf{y}}_t
    =
    \mathbf{y}_t^{w}
    +
    g_t \Delta\mathbf{y}_t .
\end{equation}
This structure keeps the refinement local and stable, because the model edits the selected
trajectory according to explicit planning feedback, while the gate controls the
magnitude of the correction.

\paragraph{Training and inference.}

For training the refiner to generate better trajectories, we construct weak/strong trajectory pair $(\mathbf{y}_t^{-},
\mathbf{y}_t^{+})$ dataset from trajectory vocabularies \citep{gtrs2025}. Given a trajectory, we sample a slightly better target trajectory from the candidate pool using two principles: improving the weakest planning metric, and taking small score-improving steps in the ranked trajectory set. This pair construction teaches the model to perform realistic local improvement, instead of learning arbitrary long-range trajectory jumps. Given a trajectory-pair $\mathbf{y}_t^{-}$, $\mathbf{y}_t^{+}$ ($\mathbf{y}_t^{-}$ plays the role of $\mathbf{y}_t^{w}$), the model is trained to match the stronger trajectory through a weighted regression objective:
\begin{equation}
    \mathcal{L}_{\mathrm{refine}}
    =
    \mathcal{L}_{\mathrm{xy}}(\bar{\mathbf{y}}_t,\mathbf{y}_t^{+})
    +
    \lambda_{h}\mathcal{L}_{\mathrm{heading}}(\bar{\mathbf{y}}_t,\mathbf{y}_t^{+}),
\end{equation}
where $\mathcal{L}_{\mathrm{xy}}$ means the L1 loss and $\mathcal{L}_{\mathrm{heading}}$ is the heading loss with the weight $\lambda_{h}$. We further randomly drop the feedback during training to enable classifier-free guidance (CFG) at inference \citep{ho2022classifier}. This reduces over-reliance on retrieved feedback and improves robustness when cross-domain memories are only partially aligned with the current scene, allowing the refiner to retain scene-conditioned correction ability. At test time, the model refines the current champion trajectory under multiple guidance strengths and outputs several corrected candidates, which are then fed back into the downstream evaluation pipeline for selection.

\subsection{Iterative Self-Evolving Inference}
\label{sec:iterative_inference}

This section ties the previous components into the test-time inference algorithm,
i.e., the concrete instantiation of the abstract reflection step $\mathcal{R}$
introduced in Sec.~\ref{sec:problem_formulation}. Each round chains SF, FAE, post-hoc feedback, and memory writing; from $n>1$ the PAR additionally consumes retrieved memory to correct the
previously selected trajectory before re-evaluation, so memory accumulation and
trajectory improvement compound across rounds.

Given $\mathcal{O}_t$, the base planner produces
$\mathcal{Y}_t^{(1)}=\pi(\mathcal{O}_t)$. At each round $n$ EvoAD applies the SF, extracts future latents for the remaining candidates, and uses the FAE to select the best proposal:
\begin{equation}
    \mathbf{y}_t^{(n)}
    =
    \arg\max_{\mathbf{y}_{t,k}^{(n)}\in\tilde{\mathcal{Y}}_t^{(n)}}
    \hat{s}_{t,k}^{(n)}
    \quad\text{with}\quad
    \hat{s}_{t,k}^{(n)} = \textsc{Eval} \bigl(\mathbf{h}_{t,k}^{(n)}\bigr).
\end{equation}
After $\mathbf{y}_t^{(n)}$ is executed, its outcome yields the feedback
$(r_t^{(n)}, \mathbf{u}_t^{(n)})$. For $n>1$, retrieved neighbours
$\mathcal{N}_t^{(n)}$ and the previous round's feedback condition the
PAR to produce a corrected proposal,
\begin{equation}
    \bar{\mathbf{y}}_t^{(n)}
    = \textsc{Refine}_{\phi}\!\left(
        \mathbf{y}_t^{(n-1)},\,\mathcal{N}_t^{(n)},\,
        r_t^{(n-1)},\,\mathbf{u}_t^{(n-1)}
      \right),
\end{equation}
which is inserted back into the candidate pool and re-evaluated alongside the
planner proposals, so the original candidates remain as fallbacks while
feedback-aware corrections are introduced progressively. The memory is updated by overwriting any earlier record of scene $t$ only when the new trajectory strictly
improves on the stored driving score. After $N$ rounds EvoAD outputs $\mathbf{y}_t^\star = \mathbf{y}_t^{(N)}$, closing a self-evolving loop in which past post-hoc reflections become priors for future refinement.

\section{Experiments}
We first compare EvoAD with representative E2E driving methods on NAVSIM~\citep{dauner2024navsim} to assess its overall planning performance. 
We then conduct module-level ablations and progressive analysis to understand where the improvement comes from and how self-evolving reflection enhances trajectory quality. 
To further examine generalization, we test EvoAD across different base planners and memory sources from different data splits. 
Finally, qualitative studies visualize how future-aware evaluation and memory-guided refinement correct unsafe or inefficient trajectories.
\subsection{Benchmark and Implementation Settings} 
We conduct experiments on NAVSIM v1 and v2~\citep{cao2025pseudo}, large-scale AD benchmarks with 100K scenarios, evaluated by PDMS and EPDMS, respectively. 
All trainable modules are trained on the \texttt{navtrain} split (85k) and evaluated on \texttt{navtest} (12k), where evaluated trajectories are accumulated as episodic memory for subsequent reflection rounds. We use WoTE~\citep{wote2025} as the base planner, which generates 256 candidate trajectories per scene.

\subsection{Main Results}

\paragraph{Comparison with baselines.}

Table~\ref{tab:main_results} compares EvoAD with existing E2E driving methods on NAVSIM \texttt{navtest}. 
Based on the WoTE planner, EvoAD (R1) first applies the SF and FAE to rerank the initial trajectory candidates, improving PDMS from 88.3 to 90.6. This shows that future-latent evaluation alone already enhances the base planner. Starting from R2, EvoAD introduces the retrieval-guided PAR, which uses previous reflection feedback to improve weak aspects of the selected trajectory. This brings a clear gain from 90.6 to 92.6 PDMS, with EP increasing from 85.7 to 88.3 and DAC from 96.4 to 97.8. 
Through self-evolution, EvoAD achieves the best performance at R5, reaching a PDMS of 93.4 on NAVSIM v1 and an EPDMS of 91.5 on NAVSIM v2. 

\begin{table}[t]
    \centering
    \caption{
        Quantitative comparison on the \texttt{navtest} split of NAVSIM v1 (PDMS) and v2 (EPDMS).
    EvoAD is built on WoTE, and R$n$ denotes the $n$-th reflection round. NC: no at-fault collision; DAC: drivable area compliance; TTC: time-to-collision; Comf.: comfort; EP: ego progress. The best results are in bold. $^{*}$Original v2 evaluation protocol.
    }
    \label{tab:main_results}
    \fontsize{8}{9.5}\selectfont
    \setlength{\tabcolsep}{3pt}
    \renewcommand{\arraystretch}{0.88}
    \begin{tabular}{l c c  c c c c c c}
        \toprule
        Method & Venue &  NC $\uparrow$ & DAC $\uparrow$ & TTC $\uparrow$ & Comf. $\uparrow$ & EP $\uparrow$ & PDMS $\uparrow$ & EPDMS $\uparrow$ \\
        \midrule
        UniAD \citep{hu2023planning}          & CVPR'23   & 97.8  & 91.9  & 92.9  & \textbf{100} & 78.8  & 83.4 & -- \\
        PARA-Drive \citep{weng2024paradrive}     & CVPR'24   & 97.9  & 92.4  & 93.0  & 99.8         & 79.3  & 84.0 & -- \\
        DRAMA \citep{yuan2024drama}         & ISRR'24   & 98.0  & 93.1  & 94.8  & \textbf{100} & 80.1  & 85.5 & -- \\
        TransFuser \citep{chitta2023transfuser}     & TPAMI'22  & 97.7  & 92.8  & 92.8  & \textbf{100} & 79.2  & 84.0 & 76.7$^{*}$ \\
        Hydra-MDP++ \citep{li2025hydra}   & arXiv'25  & 98.6  & \textbf{98.6}  & 95.1  & \textbf{100} & 85.7  & 91.0 & 85.1$^{*}$ \\
        DiffusionDrive \citep{liao2025diffusiondrive} & CVPR'25   & 98.2  & 96.2  & 94.7  & \textbf{100} & 82.2  & 88.1 & 84.8$^{*}$ \\
        WoTE  \citep{wote2025}         & ICCV'25  & 98.5  & 96.8  & 94.9  & 99.9         & 81.9  & 88.3 & 84.2 \\
        iPad  \citep{niu2025ipad}         & arXiv'25  & 98.6  & 98.3 & 94.9 & \textbf{100} & 88.0 & 91.7 & 84.1 \\
        DrivoR  \citep{sun2026drivor}       & CVPR'26   & 98.9  & 98.3 & 96.2 & \textbf{100} & 89.1 & 93.1 & 88.4 \\
        SparseDriveV2 \citep{sun2026sparsedrivev2} & ECCV'26 & 98.5 & 98.4 & 95.0 & 99.9 & 88.6 & 92.0 & 90.1 \\
        \midrule
        EvoAD (R1) & --         & 99.2 & 96.4 & 97.0 & 99.9 & 85.7 & 90.6 & 87.0 \\
        EvoAD (R2) & --         & \textbf{99.3} & 97.8 & 97.3 & 99.9 & 88.3 & 92.6 & 89.6 \\
        EvoAD (R5) & --        & \textbf{99.3} & 98.1 & \textbf{97.6} & 99.9 & \textbf{89.4} & \textbf{93.4} & \textbf{91.5} \\
        \bottomrule
    \end{tabular}
    \vspace{-1em}
\end{table}

\begin{table}[t]
    \centering
    \fontsize{8}{9.5}\selectfont

    \begin{minipage}[t]{0.49\linewidth}
        \centering
        \caption{Module-level ablation studies on NAVSIM v1 \texttt{navtest} (PDMS). \checkmark\ marks the enabled modules; WoTE and GTRS + Memory are baselines without our modules.}
        \label{tab:module_ablation}
        \vspace{-2.8pt}
        \setlength{\tabcolsep}{2pt}
        \renewcommand{\arraystretch}{0.9}
        \begin{tabular}{l c c c c c c}
            \toprule
            Method & SF & FAE & PAR & R1 & R2 & R5  \\
            \midrule
            WoTE & -- & -- & -- & 88.33 & 88.33 & 88.33  \\
            GTRS + Memory & -- & -- & -- & 90.27 & 90.35 & 90.44  \\
            \midrule
            WoTE + SF + Memory & \checkmark & -- & -- & 89.05 & 89.94 & 90.87  \\
            + only FAE & \checkmark & \checkmark & -- & 90.58 & 91.58 & 92.76  \\
            + FAE and PAR & \checkmark & \checkmark & \checkmark & \textbf{90.58} & \textbf{92.61} & \textbf{93.36}  \\
            \bottomrule
        \end{tabular}
    \end{minipage}
    \hfill
    \begin{minipage}[t]{0.49\linewidth}
        \centering
        \caption{
        Generalization studies on different base planners. Ori. denotes the original single-modality output; R2 is the mean of four refined candidates generated with four CFG strengths in Round 2.
        }
        \label{tab:cross_planner_oracle}
        \setlength{\tabcolsep}{2.4pt}
        \renewcommand{\arraystretch}{0.9}
        \begin{tabular}{l c c c c c c}
            \toprule
            Planner & Ori. & R2 
            & NC & DAC & EP & TTC  \\
            \midrule
            TransFuser 
            & 84.24 & 89.79 
            & 98.48 & 96.80 & 85.14 & 95.10  \\

            DiffusionDrive 
            & 88.19 & 91.33 
            & 98.62 & 97.86 & 86.39 & 96.09  \\

            WoTE 
            & 88.33 & 91.24 
            & 98.94 & 98.01 & 85.61 & 96.46  \\

            iPad
            & \textbf{91.91} & \textbf{94.02}
            & \textbf{99.05} & \textbf{99.13} & \textbf{90.85} & \textbf{96.38}  \\
            \bottomrule
        \end{tabular}
    \end{minipage}
    \vspace{-1em}
\end{table}

\begin{figure}[t]
    \centering
    \begin{minipage}[t]{0.36\linewidth}
        \centering
        \includegraphics[width=\linewidth]{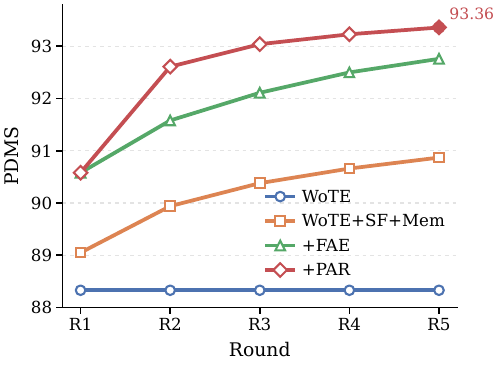}
        \caption{
        Improvement across reflection rounds.
        }
        \label{fig:progressive_rounds}
        
    \end{minipage}
    \hspace{0.1\linewidth}
    \begin{minipage}[t]{0.36\linewidth}
        \centering
        \includegraphics[width=\linewidth]{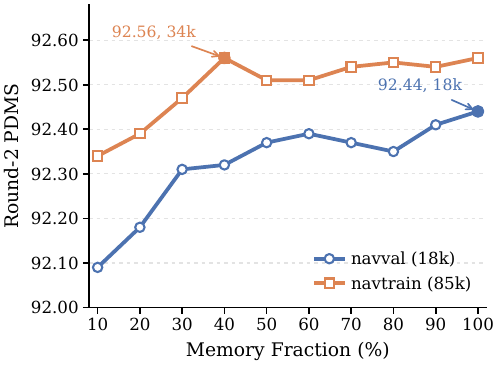}
        \caption{
        Effect of retrieved memory scale on trajectory refinement.
        }
        \label{fig:memory_fraction}
        
    \end{minipage}
\end{figure}

\paragraph{Ablation studies.}
\label{sec:ablation_studies}

Table~\ref{tab:module_ablation} reports the module-level ablation of EvoAD. 
Starting from the WoTE+SF+Memory baseline, adding FAE improves the final R5 PDMS from 90.87 to 92.76, bringing the largest gain of +1.89. 
This indicates that trajectory-conditioned future latents provide an effective reranking signal beyond safety filtering alone.
Further adding PAR improves R5 PDMS from 92.76 to 93.36, showing that feedback-guided residual refinement can further enhance the selected trajectory.

\paragraph{Multi-round self-evolution analysis.}
\label{sec:progressive_analysis}

Fig.~\ref{fig:progressive_rounds} illustrates the improvement across reflection rounds. 
Without the proposed FAE and refinement modules, the WoTE baseline remains unchanged and the WoTE+SF+Memory variant improves only gradually from 89.05 to 90.87, showing that memory alone is insufficient for self-evolution. 
The full EvoAD further improves from 90.58 at R1 to 92.61 at R2 and 93.36 at R5, demonstrating that retrieval-guided planning-aware refinement starts to take effect from the second round and brings further gains in later rounds.

\paragraph{Effectiveness on different base planners.}
Table~\ref{tab:cross_planner_oracle} shows that the PAR produces useful improvement candidates across different base planners. The mean R2 PDMS improves over the original planning trajectory, with larger gains on weaker planners, e.g., +5.55 for TransFuser and +3.14
for DiffusionDrive. The improvement remains positive for stronger planners, including
+2.91 for WoTE and +2.11 for iPad. This indicates that the PAR is not planner-specific; it captures reusable correction patterns that generalize across
different trajectory distributions.

\paragraph{Cross-domain generalization.}
\label{sec:cross_domain}

We study whether memory accumulated from other splits can transfer to \texttt{navtest}.
As shown in Fig.~\ref{fig:memory_fraction}, both \texttt{navval} and
\texttt{navtrain} memories improve R2 performance as the memory fraction increases.
The \texttt{navval} memory improves the R2 PDMS from 92.09 to 92.44 when maintaining 18k val-split experience, while the
larger \texttt{navtrain} memory reaches 92.56 at the scale of 34k samples. This suggests that EvoAD learns
transferable trajectory-feedback priors rather than relying on split-specific
memorization.

\paragraph{Reactive closed-loop evaluation.}
\label{sec:navhard}

We further evaluate EvoAD on NavHard, where surrounding agents respond to the ego vehicle. EvoAD improves across reflection rounds, with EPDMS increasing from 31.31 at R1 to 41.37 at R11.



\subsection{Qualitative Studies}

\begin{figure}[!ht]
    \centering
    \includegraphics[
        width=0.95\linewidth,
        trim={2cm 8.5cm 0.5cm 8cm},
        clip
    ]{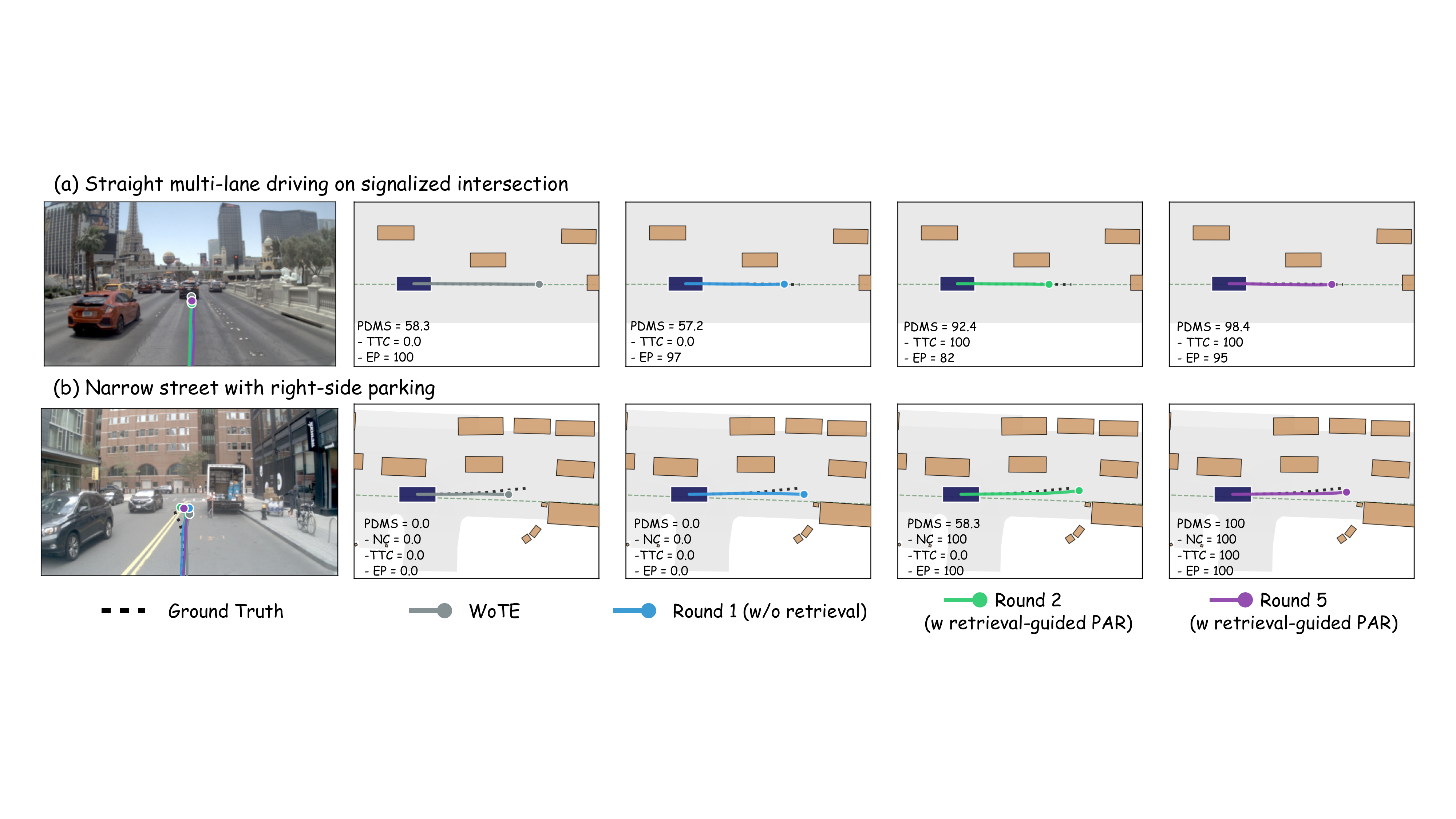}
    \caption{Qualitative improvement of EvoAD across reflection rounds. }
    \label{fig:5_improvement}

\end{figure}

\begin{figure}[!ht]
    \centering
    \includegraphics[
        width=0.95\linewidth,
        trim={2.7cm 7.5cm 1.7cm 6.9cm},
        clip
    ]{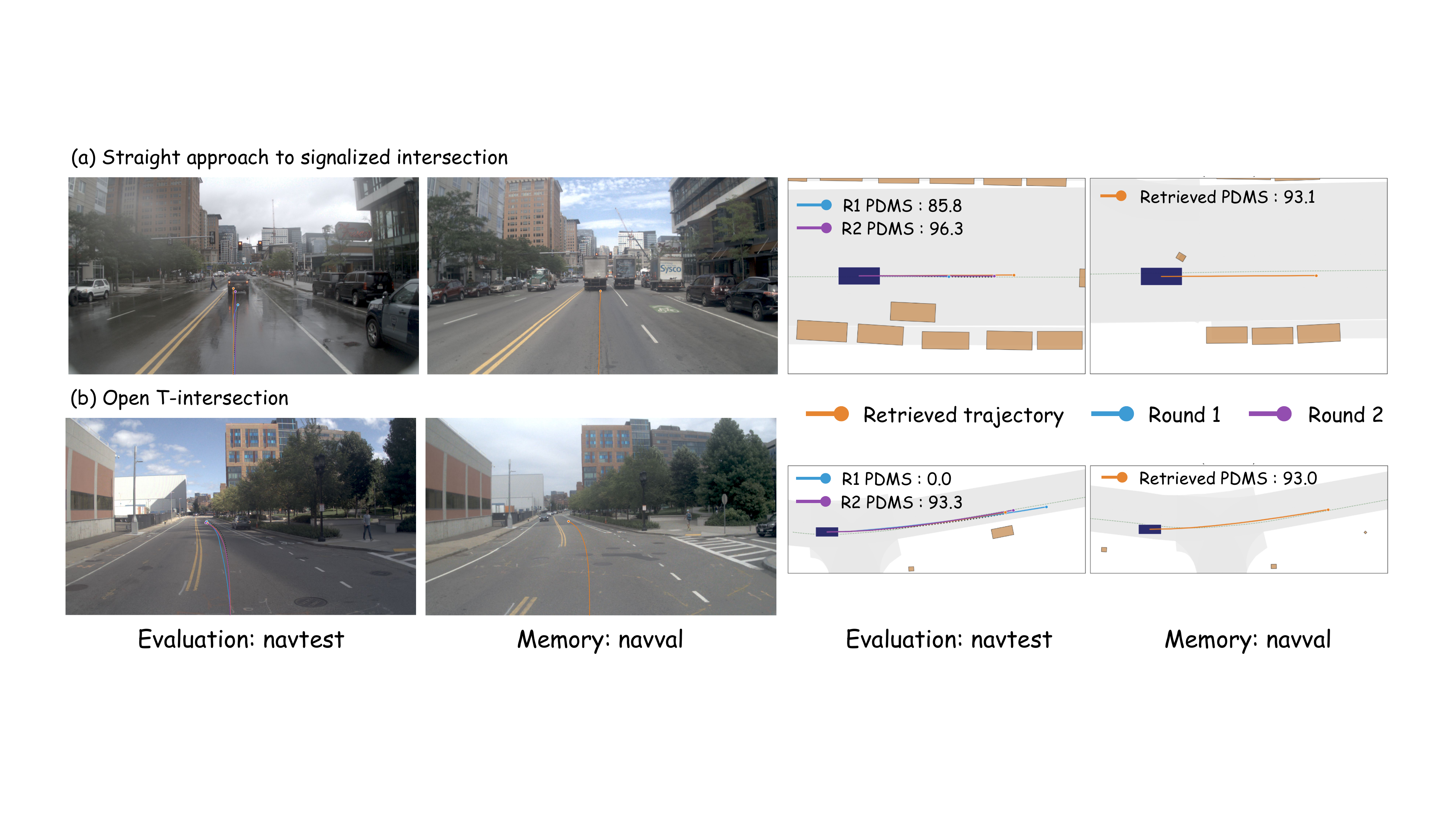}
    \caption{Generalization of memory-guided refinement across related driving scenes. }
    \label{fig:xdomain}

\end{figure}

As shown in Fig.~\ref{fig:5_improvement}, EvoAD improves trajectories through reflection rounds. 
In the signalized multi-lane intersection, reflection first removes the TTC risk and then improves efficiency, increasing PDMS from 58.3 to 98.4. 
In the narrow street with right-side parking, EvoAD corrects an initially unsafe or stalled trajectory into a collision-free and TTC-safe plan with full progress, raising PDMS from 0.0 to 100.0. Fig.~\ref{fig:xdomain} further shows that retrieved memories can transfer across related scenes. 
Even when the retrieved case differs in appearance or comes from a different split, its decision-outcome pattern provides useful refinement guidance. 
This improves the current trajectory from 85.8 to 96.3 in the signalized intersection and from 0.0 to 93.3 in the open T-intersection.



\section{Conclusion and Limitations}
We propose \textbf{EvoAD}, contributing a post-deployment self-evolution framework to E2E AD that enables vehicles to improve from past driving experiences without retraining or replacing the base model. EvoAD uses latent BEV imagination to evaluate trajectory outcomes, stores reflected experiences in memory, and retrieves relevant experiences to refine future plans. Experiments on NAVSIM v1 and v2 demonstrate the effectiveness of EvoAD, which achieves gains through multi-round reflection and generalizes across memory sources and base planners. EvoAD also has limitations. We have validated it on NAVSIM v1, v2, and the reactive NavHard split. In future work, we will extend it to more benchmarks and simulators such as CARLA~\citep{dosovitskiy2017carla}. Stronger latent world models and uncertainty-aware retrieval may further improve refinement reliability.

\bibliography{reference}
\bibliographystyle{unsrtnat}

\end{document}